\documentclass{article}
\usepackage[T1]{fontenc}
\usepackage[utf8]{inputenc}
\usepackage{libertinus}

\usepackage{macros}
\usepackage{layout}
\usepackage[authoryear,round]{natbib}
\let\parencite\citep
\let\textcite\citet

\hypersetup{pdftitle={Muon vs. GD in matrix factorization}}

\title{
Muon learns balanced solutions in matrix factorization\\
without slow saddle-to-saddle dynamics
}
\author{
Mark Rhee\eqcontrib \\ UC Berkeley
\And
Jamie Simon \\ Imbue, UC Berkeley
\And
Dhruva Karkada \\ UC Berkeley
}

\begin{document}

\makeatletter
\begingroup
    \def\Hy@footnote@currentHref{footnote.blind}
    \let\thefootnote\relax
    \footnotetext{%
    ${}^\ast$
    Correspondence to \href{mailto:mrkdh@berkeley.edu}{mrkdh@berkeley.edu}. Code is publicly available at \href{https://github.com/dkarkada/muon-mfac}{https://github.com/dkarkada/muon-mfac}.
    \parindent 21pt \par
    This work was completed during a research fellowship program at Imbue.
    }%
\endgroup
\makeatother
\maketitle
\thispagestyle{empty}

\begin{abstract}
\noindent
Matrix factorization (i.e., problems of the form $\min_{\mP,\mQ} \|\Mstar - \mP\T\mQ\|_\mathrm{F}^2$) is a minimal learning problem that exhibits both nonlinear parameter dynamics and representation learning. In this setting, we study how parameter trajectories under the Muon optimizer differ from those of gradient descent.
We identify three main dynamical differences:
\begin{enumerate}
    \item Muon avoids the slow saddle-to-saddle dynamics of GD from small initialization. Muon instead learns all the top singular vectors of $\Mstar$ at the same rate, and they saturate in order of increasing (rather than decreasing) singular value.
    \item Muon does not diverge even when the learning rate exceeds the critical threshold set by the local loss sharpness. This frees the learning rate from the condition number of the problem, enabling rapid convergence via exponential learning rate annealing.
    \item Once the weights are aligned with each other and the target, Muon flow conserves the matrix quantity $\sqrt{\mP^\top \mP}-\sqrt{\mQ^\top \mQ}$, while gradient flow is known to conserve the matrix $\mP^\top\mP - \mQ^\top\mQ$.
    Despite having distinct conserved quantities, both optimizers find the so-called \textit{balanced} solution from vanishing initialization.
\end{enumerate}
When training from small random initialization, the weights spontaneously align early in training. We derive the alignment rates in simple settings and show that they predict the empirical alignment rates in general.
Finally, we exploit structural properties of Muon to construct a learning rate schedule that achieves near-perfect alignment in only two optimization steps.

\end{abstract}

\section{Introduction}
\label{sec:intro}

\begin{figure}
\centering
\renewcommand{\arraystretch}{1.35}
\begin{tabular}{p{0.42\textwidth}p{0.42\textwidth}}
\toprule
\multicolumn{1}{c}{\textbf{Gradient Descent}} & \multicolumn{1}{c}{\textbf{Muon}} \\
\midrule
\begin{center}
    \vspace{-12pt}
    \includegraphics[width=0.75\linewidth]{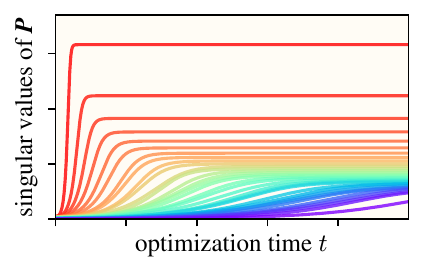}
    \vspace{-6pt}
\end{center}
\textbf{Multiplicative dynamics $\to$ slow saddle-to-saddle learning.}
Each of the weight's singular values grows at a rate proportional to its current magnitude. Singular directions slowly escape the origin one at a time. Larger modes saturate before smaller ones.
& 
\begin{center}
    \vspace{-12pt}
    \includegraphics[width=0.75\linewidth]{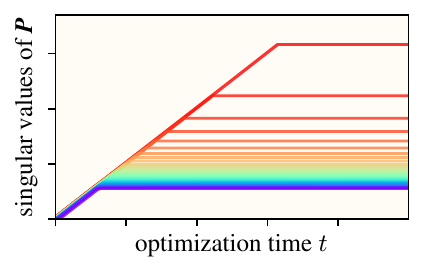}
    \vspace{-6pt}
\end{center}
\textbf{Additive dynamics $\to$ uniform growth, no intermediate saddles.}
After alignment, each singular value increases by $\eta$ per step, so all singular directions are learned at the same rate with no intermediate saddles. Thus, smaller modes saturate before larger ones. \\
\midrule
\begin{center}
    \vspace{-12pt}
    \hspace{-20pt}
    \includegraphics[width=0.85\linewidth]{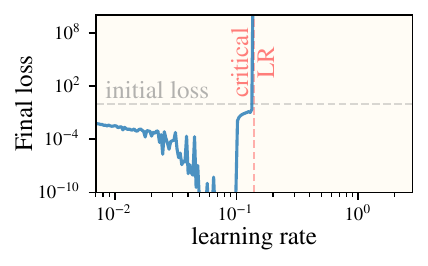}
    \vspace{-6pt}
\end{center}
\textbf{Maximum stable learning rate.} Optimization diverges if the learning rate exceeds the critical threshold $2/\lambda_{\max}(\nabla^2\mc L)$. The rate of convergence is thus limited by the condition number of the problem.
&
\begin{center}
    \vspace{-12pt}
    \hspace{-20pt}
    \includegraphics[width=0.85\linewidth]{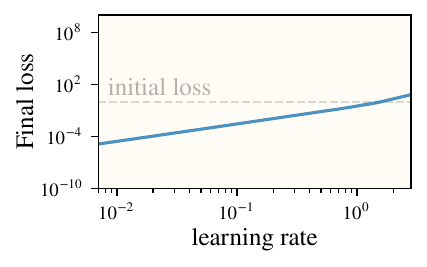}
    \vspace{-6pt}
\end{center}
\textbf{Never diverges.} 
The update magnitude is fixed and cannot diverge. With large learning rates, optimization proceeds quickly to the solution manifold and oscillates there until the learning rate is annealed. \\
\midrule
\begin{center}
    \vspace{-12pt}
    \hspace{-16pt}
    \includegraphics[width=0.82\linewidth]{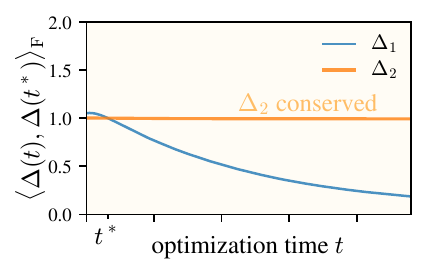}
    \vspace{-6pt}
\end{center}
\textbf{Flow lines are hyperbolic.}
The quantity $\mDelta_2\defn \mP^\top\mP-\mQ^\top\mQ$ is conserved under gradient flow. Trajectories in parameter space have hyperbolic geometry.
& 
\begin{center}
    \vspace{-12pt}
    \hspace{-16pt}
    \includegraphics[width=0.82\linewidth]{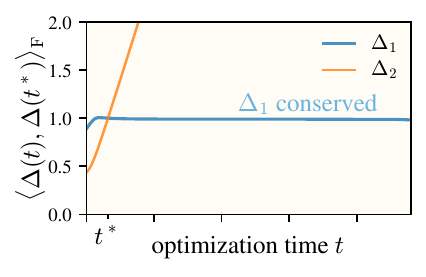}
    \vspace{-6pt}
\end{center}
\textbf{Flow lines are affine.}
After alignment, the quantity 
$\mDelta_1 \defn \sqrt{\mP^\top\mP}-\sqrt{\mQ^\top\mQ}$ 
is constant under Muon flow; the flow equations are Hamiltonian with $H(\mP,\mQ)=\mathrm{Tr}(\mDelta_1)$.
Optimization trajectories are parallel rays.
\\
\bottomrule
\end{tabular}
\caption{Muon vs. gradient descent in matrix factorization.
See \cref{app:expt} for experimental details. 
}
\label{fig:summary}
\end{figure}

Muon is a first-order optimizer which has recently found success in accelerating training in large models. Its implementation details vary; in this paper, we study the momentum-free update rule%
\begin{equation}
    \Delta \mW = -\eta \cdot \ort{\nabla_\mW\mc{L}},
    \tag{Muon}
\end{equation}
where $\mW$ is a weight matrix, $\eta$ is the learning rate, $\nabla_\mW\mc{L}$ is the gradient, and $\ort{\mG}$ denotes the \textit{orthogonalization} of $\mG$: if its singular value decomposition is $\mG = \mU \! \mS \mV^\top$, then $\ort{\mG} = \mU \mV^\top$.
Therefore, all Muon updates are (semi-)orthogonal matrices scaled by $\eta$. In contrast, the regular gradient descent update, $\Delta \mW = -\eta \cdot \nabla_\mW\mc{L}$,
directly inherits the spectrum of the gradient.

In this paper, we study the consequences of this difference for matrix factorization problems,
\begin{equation}
    \mc{L}(\mP,\mQ) = \frac{1}{2}\left\|\Mstar - \mP\mQ^\top\right\|_\mathrm{F}^2,
\end{equation}
where $\Mstar$ is a target matrix and $\mP$ and $\mQ$ are trainable weights.

Matrix factorization problems are interesting because the model $\mP\mQ^\top$ exhibits nonlinear parameter dynamics, learns nontrivial representations of the data (e.g., in word embedding algorithms or PCA), and has large-width inductive biases that illuminate those of neural networks.
In addition, matrix factorization finds use in applications such as recommender systems and LoRA fine-tuning \parencite{srebro2004learning,hu2022lora}. 

We show that, starting from small initial weights, the Muon learning dynamics unfold as follows:

\begin{enumerate}[label=(\arabic*)]
    \item At early time, the right singular vectors of $\mP$ and $\mQ$ align with each other, and the singular vectors of the model $\mP\mQ^\top$ align with those of the target. We derive the alignment rates in simple settings in \cref{sec:alignment} and find a close match with general-case empirics.

    \item As alignment completes, the singular values of each weight matrix grow uniformly. Each singular value of the model saturates when it reaches its target, oscillating \textit{ad infinitum} within a neighborhood of the target. Full convergence can only be achieved by annealing the learning rate; geometric annealing suffices. We derive these facts in \cref{sec:scalarfac}.
\end{enumerate}

The post-alignment behaviors in particular are qualitatively different from gradient descent, which exhibits stepwise learning from small initialization and whose convergence is sensitive to the learning rate (see \cref{fig:summary}). However, similar to GD, as the initialization scale vanishes Muon reaches increasingly ``balanced'' solutions (i.e., the final weight matrices are increasingly aligned and commensurate in norm). This is because the duration of the alignment phase (where imbalance accumulates) shortens with vanishing random initialization.

In \cref{sec:schedule}, we design a learning rate schedule that achieves alignment in only two optimization steps, in the case where the model has enough parameters to completely learn the target. The schedule consists of three stages: a small initial step that aligns $\mP$ with $\mQ$, followed by a large step that simultaneously aligns the model with the target and drives it to the solution manifold, and finally an annealing stage during which the model converges to a solution.

Though we do not rigorously prove these claims, we support all of them with quantitative calculations and empirical evidence. Our goal is to provide simple yet substantiated intuition about the behavior of Muon---in particular, that its learning dynamics quickly guide the model to desirable (balanced) solutions.

\subsection{Setup and notation}

We use capital boldface to denote matrices and lowercase boldface for vectors.
Parenthesized subscripts denote tensor elements (e.g., $\mA_\idx{ij}$ is a scalar).
Parenthesized superscripts denote optimization iterates (e.g., $\mP^\idx{t}$ is $\mP$ after $t$ steps); we leave off the superscript when it is unimportant or clear from context.
For a matrix $\mA$, $\|\mA\|_F$ denotes the Frobenius norm, equal to the square root of the sum of squared singular values (equivalently, of squared entries); $\|\mA\|_\ast$ denotes the nuclear norm, the sum of the singular values.

The learning problem is $\mc{L}(\mP,\mQ) = \frac{1}{2}\left\|\Mstar - \mP\mQ^\top\right\|^2$. $\Mstar\in\R^{n\times n}$ is a fixed square target matrix.%
\footnote{We focus on square $\Mstar$ for simplicity; our analysis extends straightforwardly to non-square targets.}
$\mP\in\R^{n\times d}$ and $\mQ\in\R^{n\times d}$ are both trainable weights whose discrete dynamics are driven by the Muon iteration rule with learning rate $\eta$:
\begin{align}
    \Delta \mP &= -\eta \cdot \ort{\nabla_\mP \mc L} = \eta \cdot \ort{\mR\mQ} \\
    \Delta \mQ &= -\eta \cdot \ort{\nabla_\mQ \mc L} = \eta \cdot \ort{\mR^\top\mP},
\end{align}
where we define the (time-dependent) residual as $\mR\defn \Mstar - \mP\mQ^\top$.
We use the wide hat to denote the matrix sign function, i.e., polar orthogonalization:%
\footnote{If $\mG$ is a vector, then matrix sign reduces to unit-normalization, justifying the hat notation. We do not use hats to denote statistical estimates (since our problem, as framed, does not involve statistical estimation).}
\begin{equation}
    \ort{ \mG} \defn \mathrm{msign}(\mG) = \mU\mV^\top \quad\text{where } \mG=\mU\mS\T\mV.
\end{equation}

The initial conditions are sampled from the isotropic Gaussian: $\mP_\idx{ij}^\idx{0} \sim\mc{N}(0,\frac{\alpha^2}{{\max(n,d)}})$ and likewise for $\mQ$. The initialization variance depends on whether the model is over-parameterized ($d\geq n$) or under-parameterized ($d<n$), because we want the relative size of the update to scale only with the initialization scale $\alpha$ and the learning rate $\eta$, and not with the model dimensions:
\begin{equation}
    \frac{\|\Delta \mP\|^2}{\|\mP^{(0)}\|^2} \approx \frac{\|\Delta \mQ\|^2}{\|\mQ^{(0)}\|^2}
    \sim \frac{\eta^2 \min(n,d)}{(\alpha^2/\max(n,d)) \cdot nd}
    = \frac{\eta^2}{\alpha^2}
\end{equation}
using standard concentration arguments.%
\footnote{We will later find that $\|\mP^\idx{t}\|$ grows linearly with $t$, so one can warm up the learning rate linearly and still ensure $\|\Delta \mP\|/\|\mP\| \ll 1$ throughout the growth phase, i.e., that the trajectory is well-described by a continuous flow.}
This convention is consistent with prior literature using Muon at scale \parencite{liu2025muon}; however, this argument privileges the Frobenius norm, and is therefore inconsistent with $\mu P$, where the native norm is the spectral norm \parencite{qiu2026hyperparameter,yang2023spectral}. (We are not constrained by the $\mu P$ scaling rule since we are not taking the $d/n\to\infty$ limit.)

To understand the dynamics of singular vector alignment in this flow-like regime, we define the following SVDs:
\begin{equation}
    \Mstar=\mU^\ast\mS^\ast{\mV^\ast}^\top \qquad\qquad \mP\mQ^\top = \mU_\mathrm{mod}\mS_\mathrm{mod} \mV_\mathrm{mod}^\top
\end{equation}
and
\begin{equation}
    \mP=\mU_P\mS_P\mV_P^\top \qquad\qquad \mQ=\mU_Q\mS_Q\mV_Q^\top.
\end{equation}
Intuitively, the alignment between the weights is quantified by the overlap%
\footnote{See \cref{sec:alignment} for precise definitions of alignment and overlap.}
between $\mV_P$ and $\mV_Q$, and the model-target alignment is quantified by the overlap between $\mU_\mathrm{mod}$ and $\mU^\ast$, and between $\mV_\mathrm{mod}$ and $\mV^\ast$. 
If these singular vectors are fully aligned, and if $\mS_P \approx \mS_Q$ (which is the case when the initialization is small, see \cref{sec:scalarfac}), then the optimization trajectory will approach the set of \textit{balanced solutions} $\{\mP,\mQ: \mP\mQ^\top = \mathrm{top}_d(\Mstar), \mP^\top\mP = \mQ^\top\mQ\}$.

Without loss of generality, we will always assume that $\Mstar$ is diagonal with nonnegative entries. We do this by simply rotating the whole problem into the SVD basis of $\Mstar$. Concretely, define the rotated weight matrices $\tilde{\mP} := {\mU^{*}}^\top \mP$ and $\tilde{\mQ} := {\mV^{*}}^\top \mQ$. Then, the loss becomes $\tilde{\mc L}=\|\mS^* - \tilde{\mP} \tilde{\mQ}^\top \|_F^2$ and the update rule becomes
\begin{align*}
    \Delta \tilde{\mP} &= -\eta \cdot \ort{\nabla_{\tilde{\mP}} \tilde{\mc L}} = \eta \cdot \ort{\tilde{\mR}\tilde{\mQ}} \\
    \Delta \tilde{\mQ} &= -\eta \cdot \ort{\nabla_{\tilde{\mQ}} \tilde{\mc L}} = \eta \cdot \ort{\tilde{\mR}^\top\tilde{\mP}},
\end{align*}
where $\tilde{\mR} = \mS^* - \tilde{\mP}\tilde{\mQ}^\top$ is the rotated residual. Since the equations are the same in the new variables, and since the initialization distributions of $\tilde\mP$ and $\tilde\mQ$ are the same as those of $\mP$ and $\mQ$, we can simply drop the tildes to work directly in the diagonal basis of $\Mstar$.

\section{Related Work}
\label{sec:related}

\paragraph{Muon in matrix factorization.}
Closest to our work, \textcite{ma2026} and \textcite{kang2026} analyze Muon in the matrix factorization setting. Both identify a \emph{uniform growth} phase in which every singular value increases at the same rate, giving linear convergence whose iteration complexity is independent of the target's condition number. Both works rigorously prove convergence rates using techniques that avoid directly characterizing the alignment phase. We instead pursue a mechanistic understanding of alignment and growth via scaling arguments. Our analysis, while less rigorous, is intuitive and correctly predicts the empirical alignment dynamics in general settings.

\paragraph{Learning dynamics of gradient descent in matrix factorization.}
Matrix factorization is a standard problem setting to study deep learning phenomena. 
From small initialization, gradient descent on a matrix factorization objective has a pronounced low-rank bias, greedily recovering the target one effective rank at a time via sigmoidal parameter trajectories \parencite{saxe2014,gunasekar2017, arora2019, gidel2019, li2021greedy}. 
The same objective is also a minimal model of representation learning: \textcite{karkada2025} show that a matrix factorization objective correctly predicts the learning dynamics of the \texttt{word2vec} word embedding algorithm.

A recurring theme is that training from small initialization separates into an early \emph{alignment} phase and a later \emph{growth} phase. \textcite{atanasov2022} term this \emph{silent alignment}: the singular vectors of the model align with those of the target while the loss is still essentially flat, on a timescale governed by the ratio between the target scale and the initialization scale, after which the singular values grow. Asymptotically, gradient descent provably aligns the factors both with each other and with the target \parencite{ji2019, radhakrishnan2020}, and \textcite{stoger2021} show that this alignment sets in early: from small random initialization, a few gradient steps already place the factors near the top singular subspace of the target, so the remainder of training proceeds as though it had been spectrally initialized. The growth phase itself proceeds as \emph{saddle-to-saddle} dynamics: trajectories linger near a sequence of low-rank saddles and switch on one mode at a time, producing the characteristic staircase loss curve \parencite{jacot2021}. These dynamics are organized by a conservation law, namely that the difference of squared factor norms is invariant under gradient flow, so that small initialization keeps the two factors balanced throughout training \parencite{du2018, marcotte2023}; \textcite{soltanolkotabi2023} combine this implicit balancing with the early alignment above to obtain convergence and generalization guarantees for asymmetric matrix sensing, where the target is observed only through random linear measurements rather than directly. Our analysis revisits each of these three phenomena---alignment, growth, and conservation---under Muon.

\paragraph{When Muon outperforms gradient descent.} Muon is an optimizer that flattens the spectrum of matrix-valued updates \parencite{jordan2024}. It has driven strong empirical results \parencite{liu2025muon}. \textcite{bernstein2024} cast this update as steepest descent under the spectral norm. A consistent picture across works that compare Muon with gradient descent is that Muon helps precisely when the problem carries heterogeneous spectral structure---a low-rank, ill-conditioned, or imbalanced component that GD learns slowly. This mechanism has been made precise for class-imbalanced classification, where spectral descent and Muon learn all principal components at equal rates and thereby recover the tail classes that GD and Adam learn slowly \parencite{vasudeva2025, wang2025tail, fan2025, li2026assoc}, and for deep networks more broadly, where low-stable-rank activations induce an ill-conditioned Euclidean landscape to which spectral updates are naturally adapted \parencite{davis2025, braun2026}. Ill-conditioned matrix factorization is the same
phenomenon in a pure optimization setting, with the spread of target singular values playing the role of the imbalance ratio.

\section{Learning dynamics after weight alignment}
\label{sec:scalarfac}

We will first assume that the weights are already fully aligned, i.e., that the model is aligned with the target ($\mP\mQ^\top$ commutes with $\Mstar$) and the weights are aligned with each other ($\mP^\top\mP$ commutes with $\mQ^\top\mQ$).
Then the residual $\mR = \Mstar - \mP\T\mQ$ is diagonal in the target's SVD basis, and the learning problem decouples into $n$ independent scalar problems \parencite{ma2026}:
\begin{align}
    \mc{L} &= \frac{1}{2}\|\mR\|^2 \\
    &= \frac{1}{2}\sum_{\mu=1}^n (m^\ast_\mu - p_\mu q_\mu)^2
\end{align}
where $m^\ast_\mu$, $p_\mu$, $q_\mu$ are the $\mu$-th largest singular values of $\Mstar$, $\mP$, $\mQ$ respectively. Without loss of generality, we can directly optimize $p_\mu$ and $q_\mu$, due to the model's internal rotation symmetry (i.e., neither the task nor the learning rule prefers a particular basis in the model's hidden latent space).

\begin{figure}[t!]
    \centering
    \includegraphics[width=0.8\linewidth]{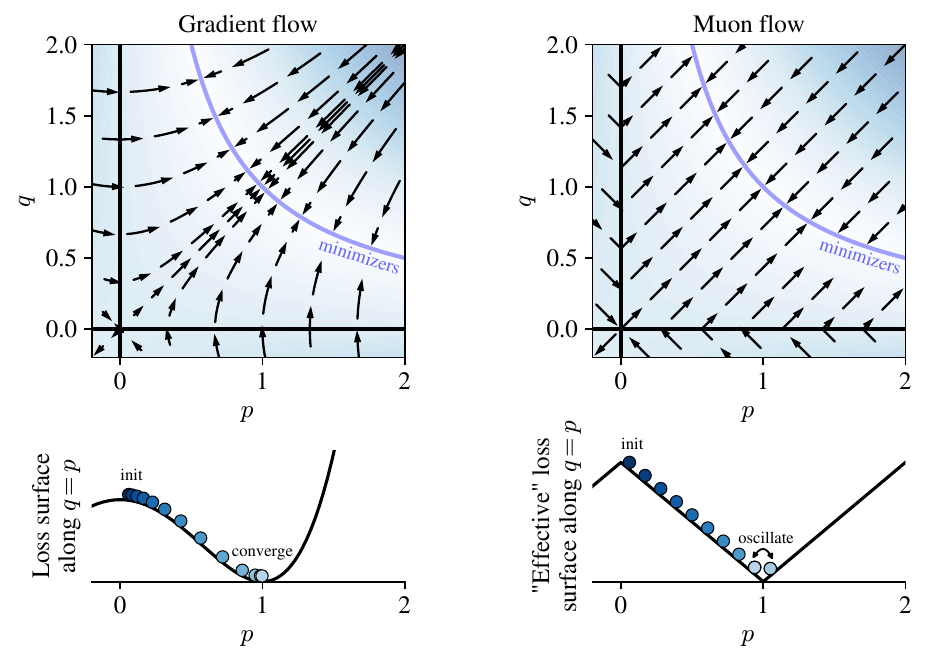}
    \caption{
        \textbf{In GD, flow lines are hyperbolic with slow saddle points; in Muon, they are parallel and uniform.}
        \textbf{(Top.)} We show optimization flow trajectories in parameter space for scalar factorization, $\mathcal{L}(p,q) = \frac{1}{2}(1 - pq)^2$, comparing gradient descent (left) and Muon (right). Arrow lengths indicate update magnitude, and the solution manifold $pq=1$ is shown in blue.
        \textbf{(Bottom.)} We show a slice of the (effective) loss landscape along the line $p=q$. The ``effective loss'' $\mc L_\mathrm{eff}$ is defined such that the GD trajectory on $\mc L_\mathrm{eff}$ gives the same trajectory as Muon on $\mc L$.
        In GD, small initialization is cursed by slow convergence due to the saddle point at the origin. Muon escapes this curse since its flow rate is uniform arbitrarily close to the origin. However, unlike GD, Muon with constant learning rate oscillates about the solution manifold without converging, so it is necessary to anneal the learning rate.
    }
    \label{fig:fig2}
\end{figure}

Each of these scalar problems minimizes the loss $\mc{L}(p,q) = \frac{1}{2}(m^\ast -pq)^2$. The Muon update is 
\begin{align}
    \Delta p &= \eta\cdot \ort{rq}= \eta\cdot\mathrm{sign}(r) \\
    \Delta q &= \eta\cdot \ort{rp} = \eta\cdot\mathrm{sign}(r)
\end{align}
where $r \defn m^\ast -pq$ is the residual, and we assumed nonzero singular values, i.e., $p,q>0$. Already, we see some striking differences from gradient descent (see \cref{fig:fig2} for a visualization):
\begin{itemize}
    \item The magnitude of the update is fixed. This implies that each singular value flows at the same uniform rate, with no dependence on its current value or the target value. This sharply contrasts gradient descent, where singular values grow exponentially, separating from each other in the small initialization regime, leading to stepwise learning.
    \item For fixed $\eta$, the iterates can neither diverge nor converge. The ``effective sharpness'' is zero everywhere except on the measure-zero set where the gradient is ill-defined, so the loss cannot diverge regardless of the learning rate. On the other hand, the trajectory does not slow down near the minimizer, causing the iterates to oscillate across the solution manifold due to the changing $\sgn(r)$.
    \item When training from small initialization, $r>0$ before reaching the solution manifold. In the flow limit, $\dot{p}=\dot{q}=1$, and $p-q$ is a conserved quantity.
\end{itemize}

Another way to understand the last observation is to import the following fact from classical mechanics: a flow $(\dot{p}, \dot{q})$ is Hamiltonian if it is generated by some scalar $H(p,q)$ via $\dot{q} = \partial H/\partial p$ and $\dot{p} = -\partial H/\partial q$. The $r>0$ Muon flow is generated by the Hamiltonian $H(p,q) = p-q$. Hamiltonian flows have the property that $H$ is an invariant of the flow, and is therefore a conserved quantity. The orbits of this Hamiltonian are parallel in parameter space; this contrasts gradient flow, whose flow lines are hyperbolas (since gradient flow conserves the quantity $p^2-q^2$, through a different, non-Hamiltonian mechanism).

It follows that the matrix generalization of this conserved quantity is $\mDelta_1 \defn \sqrt{\mP^\top \mP}-\sqrt{\mQ^\top \mQ}$. Likewise, the flow-generating Hamiltonian is $H(\mP,\mQ)=\Tr \mDelta_1=\|\mP\|_\ast - \|\mQ\|_\ast$, where $\|\cdot\|_\ast$ is the nuclear norm. If the weights are aligned and $\mDelta_1 \approx 0$, then Muon will reach a (nearly-)balanced solution.

We note that the logical relation between the conservation law and alignment is different from GD. In GD, the conserved quantity holds at all times and is a consequence of the learning rule and symmetry of the problem, and can thus be used to argue that the weights must quickly align. In Muon, however, we assume alignment as a precondition and use it to derive a conservation law; the conservation law \textit{does not} hold before alignment is achieved. Clearly, the dynamics of alignment are important to understand; we study them in detail in \cref{sec:alignment}.

\subsection{Oscillations near the solution manifold}
\label{sec:anneal}

With constant learning rate, the Muon iterates oscillate about the solution manifold without converging. To converge, then, it is necessary to anneal the learning rate. In particular, geometrically annealing the learning rate by a factor of $1/2$ at each step gives linear convergence. We give a direct proof here in the case of aligned weights near the solution manifold, and refer to \textcite{ma2026} for a more general analysis.

The conservation law ensures that the parameters stay balanced. Assume as the inductive hypothesis that at time $t$, $|p^{(t)}-\sqrt{m^*}|<\eta^{(t-1)}$. Without loss of generality, assume $p^{(t)}-\sqrt{m^*} < 0$ (the $p^{(t)}-\sqrt{m^*} > 0$ case holds by a symmetric argument). Then $p^{(t+1)} = p^{(t)}+\eta^{(t)}$ where $\eta^{(t)}=\frac{\eta^{(t-1)}}{2}$. The inductive step maintains the inductive hypothesis as follows: 
\begin{align*}
    \left | p^{(t+1)}-\sqrt{m^*} \right | &= \left |p^{(t)}-\sqrt{m^*}+\frac{\eta^{(t-1)}}{2} \right |\\
    &< \left |-\eta^{(t-1)}+\frac{\eta^{(t-1)}}{2} \right | \\
    &= \frac{\eta^{(t-1)}}{2} = \eta^{(t)}.
\end{align*}
The convergence rate of $p^{(t)}\to \sqrt{m^*}$ is thus upper bounded by the geometric decay of $\eta^{(t)}$.

\subsection{Oscillation traps near the origin}

The solution manifold is not the only place where the iterates can oscillate. If the initial weights are very misaligned and the learning rate is too large, the iterates can oscillate across the origin for a long time.
One can understand this from the flow map in \cref{fig:fig2} by considering the initialization $(p_0, q_0)=(-\alpha, \alpha)$ with $\eta=\alpha$; the iterates oscillate indefinitely between the second and fourth quadrants.
In both the scalar and matrix cases, these oscillation traps can be avoided by going deeper into the flow regime (i.e., decreasing $\eta/\alpha$).
In the matrix case, this trap can also be avoided by heavy overparameterization, which makes catastrophic misalignment much less likely.

\section{Why we can expect model subspaces to align quickly}
\label{sec:alignment}

In matrix factorization problems, the most important aspects of learning are contained in the alignment dynamics. It is during alignment that the model learns to directly encode the most important target directions within the singular directions of the weights themselves.
This nontrivial behavior is key to learning task-relevant features; over-parameterized models in the neural tangent kernel regime \textit{do not} align, despite ultimately solving the matrix factorization problem with zero loss.
Therefore, it is important to understand the mechanism of rapid spontaneous weight alignment when training from small random initialization.

We begin by formally defining alignment, a scalar quantity associated to a pair of positive semi-definite matrices (with eigendecompositions $\mV\mS\mV^\top$ and $\widetilde\mV\widetilde\mS\widetilde\mV^\top$) that quantifies the extent to which they share a jointly-ordered eigenbasis. For example, when both spectra are non-degenerate, the two matrices are fully aligned iff $\mV^\top \widetilde\mV$ is diagonal with entries $\pm 1$.
In this case, we can define the scalar alignment metric $a_\mathrm{nondegen} \in [0, 1]$:
\begin{equation}
    a_\mathrm{nondegen} \defn \frac{1}{n}\sum_{\mu=1}^n\Tr[(\vv_\mu \vv_\mu^\top)(\widetilde\vv_\mu \widetilde\vv_\mu^\top)]
    = \frac{1}{n}\sum_{\mu=1}^n \cos^2(\vv_\mu,\widetilde\vv_\mu).
\end{equation}
Thus, the trace of the product of projectors is simply the squared cosine similarity, a natural and intuitive measure of overlap.

\begin{figure}[t!]
    \centering
    \includegraphics[width=1.0\linewidth]{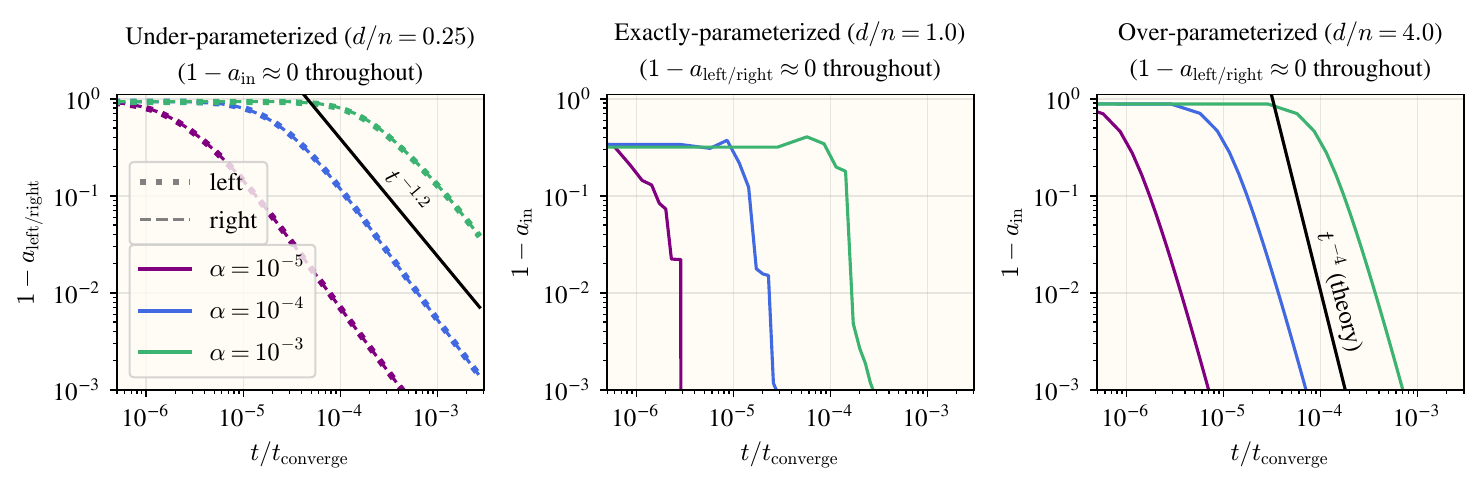}
    \caption{
    \textbf{Alignment rates vary with aspect ratio $d/n$ and occur earlier at smaller initialization scale.}
    We show the internal alignment metric $a_\mathrm{in}$ and the left and right target alignment metrics $a_\mathrm{left}$ and $a_\mathrm{right}$ for matrix factorization problem ($n=128, d\in[32,128,512]$). The target has planted low-rank-signal-in-noise structure: $\Mstar = \mG + \lambda \mU \mV^\top$, where $\mG$ is random Gaussian, $\lambda$ is the planted signal strength, and $\mU,\mV\in\R^{n\times 32}$ are semi-orthogonal. See \cref{app:expt} for details and analysis for more general target spectra.
    }
    \label{fig:fig3}
\end{figure}

However, we must generalize this notion for matrices with (nearly) degenerate subspaces. This is because the singular vectors spanning such subspaces are unstable under perturbation. To ensure that our alignment metric is not sensitive to such instabilities, we define
\begin{align}
    a(\mV,\widetilde\mV,\mS,\widetilde\mS) &\defn \frac{1}{n'}\sum_{\mu=1}^{n'}\frac{1}{\mathrm{dim}(\mu)}\Tr[(\mV_\mu \mV_\mu^\top)(\widetilde\mV_\mu \widetilde\mV_\mu^\top)]
    \\
    &\;= \frac{1}{n'}\sum_{\mu=1}^{n'} \frac{1}{\mathrm{dim}(\mu)} \left\|\mV_\mu^\top \widetilde\mV_\mu\right\|_\mathrm{F}^2,
\end{align}
where $\mu$ enumerates the $n'$ nearly-degenerate subspaces of the PSD matrices, and $\mV_\mu\in\R^{n\times\dim(\mu)}$ is the submatrix of $\mV$ containing the $\mu^\mathrm{th}$ nearly-degenerate subspace. It remains to choose how to demarcate these subspaces jointly between the two matrices, especially when their spectra differ.
In our experiments, we use a convention in which indices $i$ and $j$ are grouped in a degenerate subspace if \textit{either} $s_i/s_j \approx 1$ or $\tilde s_i/\tilde s_j \approx 1$ (see \cref{app:expt}).

With a quantitative alignment metric in hand, we may study the following alignment quantities:
\begin{alignat}{2}
    \textbf{Internal alignment.}& \qquad \qquad a_\mathrm{in} &&\defn a(\mV_P, \mV_Q, \mS_P, \mS_Q) \\
    \textbf{Left target alignment.}&\qquad  \qquad a_\mathrm{left} &&\defn a(\mU_\mathrm{mod}, \mU^\ast, \mS_\mathrm{mod}, \mS^\ast) \\
    \textbf{Right target alignment.}&\qquad \qquad a_\mathrm{right} &&\defn a(\mV_\mathrm{mod}, \mV^\ast, \mS_\mathrm{mod}, \mS^\ast).
\end{alignat}

In both our theory and experiments, we will be interested in the \textit{mis}alignment quantity $1-a$ (which, in some sense, represents the average $\sin^2(\theta)$ between subspaces). Misalignment typically starts at some $\mc O(1)$ quantity and rapidly decays to zero over the course of optimization. Our goal is to characterize this decay rate.

This analysis requires working in cases; the dynamics of internal alignment are qualitatively distinct from those of target alignment. We predict the following ``heuristic'' alignment rates by studying toy models:
\begin{itemize}
    \item In the under-parameterized regime ($d < n$), internal alignment comes essentially for free, while target misalignment decays via power iteration like
    \begin{equation}
        1-a_\mathrm{left} \approx 1-a_\mathrm{right} \sim (t/\alpha_\mathrm{eff})^{-2(1-s^{-1})},
    \end{equation}
    where $t$ is training time, $\alpha_\mathrm{eff}\defn \max(\eta, \alpha)$ is the effective%
    \footnote{If the step size $\eta$ is larger than the true initialization scale $\alpha$, then the effective initialization scale is $\eta$, since the first step will take the model to that scale.} initialization scale,
    and $s\geq1$ is the relative spectral gap in the target.
    We derive this rate theoretically in the $n=2$, $d=1$ problem $\mc{L}(\vp) = \frac{1}{2}\|\diag([s,1])-\vp\T\vp\|^2$.
    \item In the over-parameterized regime ($d > n$), target alignment comes essentially for free, while internal misalignment decays like
    \begin{equation}
        1-a_\mathrm{in} \sim (t/\alpha_\mathrm{eff})^{-4}.
    \end{equation}
    We derive this rate theoretically in the problem $\mc{L}(\vp,\vq) = \frac{1}{2}(m^\ast-\T\vp\vq)^2$.
    \item When $d=n$, we empirically find that both types of alignment happen quickly.
\end{itemize}

In sum, 
target misalignment decays polynomially with exponent at most $-2$, with rates slowing as the spectral gap in the target gets smaller.
Internal misalignment decays even more quickly: polynomially with exponent $-4$.
In all cases, alignment occurs earlier with smaller effective initialization scale. These results are summarized in \cref{fig:fig3}.

\subsection{A simple model of under-parameterized alignment}
\label{sec:outer-prod}

To study model-target alignment, we analyze the ``outer product factorization'' problem: minimize $\mc{L}(\vp) = \frac{1}{2}\|\Mstar - \vp\T{\vp}\|^2$, where $\Mstar = \diag(s, 1)$ with $s \geq 1$. Since both factors share the same parameters (tied vector weights), internal alignment is trivially satisfied; the target misalignment is $\eps \defn 1-a_\mathrm{right} = 1-(\hat{\vp}^\top\hat{e})^2$, where $\hat{e} \defn (1,0)^\top$ is the dominant eigenvector of $\Mstar$.

From small initialization, $\|\vp\|^2 \ll 1$ for much of optimization, and the residual is dominated by the target.
Therefore the Muon flow is approximately
\begin{equation}
  \dot\vp \approx \frac{\Mstar\vp}{\|\Mstar\vp\|}.
  \label{eq:normalized-flow}
\end{equation}
Absorbing the normalization into a ``rescaled clock'' via $\dd t = \|\Mstar\vp\|\,\dd\tau$ gives $\dd\vp/\dd\tau = \Mstar\vp$. With the special symmetric initialization $\vp(0) = [\alpha,\alpha]$, this integrates to
\begin{equation}
  \vp(\tau) = \alpha\begin{pmatrix}e^{s\tau}\\e^{\tau}\end{pmatrix},
  \qquad \|\Mstar\vp(\tau)\| = \alpha\sqrt{s^2e^{2s\tau}+e^{2\tau}}.
\end{equation}

Using this, we may return to the ``regular clock'' via $t(\tau) = \alpha\int_0^\tau\sqrt{s^2e^{2s\tau'}+e^{2\tau'}}\,\dd\tau'$. For non-degenerate targets (i.e., $s>1$) the first term dominates:
\begin{equation}
  t(\tau)\sim\alpha\,e^{s\tau}\quad\Longrightarrow\quad\tau(t)\sim\tfrac{1}{s}\ln(t/\alpha).
\end{equation}
Substituting back,
\begin{equation}
  \vp_\idx{0}(t)\sim t,\qquad\vp_\idx{1}(t)\sim\alpha(t/\alpha)^{1/s}.
\end{equation}

The target misalignment is then
\begin{equation}
  \eps(t) = \frac{\vp_\idx{1}^2}{\vp_\idx{0}^2+\vp_\idx{1}^2} = \frac{1}{1+\left({\vp_\idx{0}}/{\vp_\idx{1}}\right)^2} \sim \frac{1}{1+(t/\alpha)^{2(1-s^{-1})}}.
\end{equation}
Discretizing the flow with step size $\eta$, we get two qualitative regimes. When $\eta/\alpha \ll 1$ we are near the flow regime, and the above analysis holds. However, when $\eta/\alpha \gg 1$, the first step increases the parameter norms from $\alpha$ to $\approx\eta$ while barely changing the alignment. Therefore, the effective initialization scale is $\alpha_\mathrm{eff}\defn\max(\eta, \alpha)$, and the misalignment decays asymptotically like
\begin{equation}
  \boxed{\;1-a_\mathrm{right}\;\sim\;\left(\frac{t}{\max(\eta,\alpha)}\right)^{-2(1-s^{-1})}.\;}
\end{equation}
At best, when $s\to\infty$, we have $\eps\sim(t/\alpha_\mathrm{eff})^{-2}$ . At worst, when $s\to1^+$, alignment stalls completely.

\subsection{A simple model of over-parameterized alignment}
\label{sec:inner-prod}

To study internal alignment, we analyze the ``inner product factorization'' problem: minimize $\mathcal{L}(\vp,\vq) = \frac{1}{2}(m^\ast -\vp^\top \vq)^2$. Since the target $m^\ast$ is scalar, the model $\vp^\top \vq$ is always aligned to it; the misalignment between the vectors $\vp$ and $\vq$ is $\eps \defn 1-a_\mathrm{in} =  1-(\hat\vp^\top\hat\vq)^2=\sin^2(\vp,\vq)$.

With small initialization, the early-time residual $r \defn m^\ast - \vp^\top\vq$ is positive. The Muon flow equations are
\begin{equation}
  \dot\vp = \ort{\vq},\qquad \dot\vq = \ort{\vp}.
\end{equation}
These are Hamilton's equations for $H \defn \|\vp\|-\|\vq\|$. Assume balanced initialization $\|\vp(0)\|=\|\vq(0)\|=\alpha$; then $H\equiv 0$. We may define the radial variable $x\defn\|\vp\|=\|\vq\|$, whose evolution is $\dot x = \ort{\vp}^\top\ort{\vq} = \sqrt{1-\eps}$.
The Hamiltonian is invariant under $(\vp,\vq)\mapsto(\mR\vp,\mR\vq)$, so Noether's theorem gives the conserved angular momentum
\begin{equation}
  L^2 \defn \|\vp\|^2\|\vq\|^2-(\vp^\top\vq)^2 = x^4\eps = \mathrm{const}.
\end{equation}
Differentiating $x^4\eps = L^2$ and using $\dot x = \sqrt{1-\eps}$ gives $\dot\eps = -4\eps\sqrt{1-\eps}/x$. Eliminating $x$ using the angular momentum conservation law gives
\begin{equation}
  \dot\eps = -\eps^{5/4} \; \frac{4\sqrt{1-\eps}}{\sqrt{L}}.
\end{equation}
Separating variables yields an elliptic integral, but we can extract the asymptotic alignment behavior by taking $\eps\to0$. This gives
\begin{equation}
  \eps(t) \;=\; 1-a_\mathrm{in}(t) \;\sim\; \left(\frac{t}{\sqrt{L}}\right)^{-4}.
\end{equation}
At initialization $L^2 = \alpha^4\eps_0\approx\alpha^4$ (since $\eps_0 = O(1)$ for rotation-invariant init in $d\geq 2$), so $\sqrt{L}\sim\alpha$ and the flow predicts $1-a_\mathrm{in}\sim(t/\alpha)^{-4}$.
Discretizing the flow with step size $\eta$ as in the under-parameterized setting, we get
\begin{equation}
  \boxed{\;1-a_\mathrm{in}\;\sim\;\left(\frac{t}{\max(\eta,\alpha)}\right)^{-4}.}
\end{equation}
Though this rate holds at large $t$, the absolute misalignment can drop dramatically (almost to zero) in the first two steps, if $\eta\gg\alpha$. We discuss this trick in the next section.

\section{The spiked learning rate schedule}
\label{sec:schedule}

Using the theoretical insight we gained from \cref{sec:scalarfac} and \cref{sec:alignment}, we construct a learning rate schedule that achieves near-perfect alignment in two steps, in the exactly- and over-parameterized regimes. We do not claim that such a schedule is advisable in general settings; instead, it is a proof-of-concept that theoretical insight about learning dynamics can inform practice.

\begin{figure}[t!]
    \centering
    \includegraphics[width=1.0\linewidth]{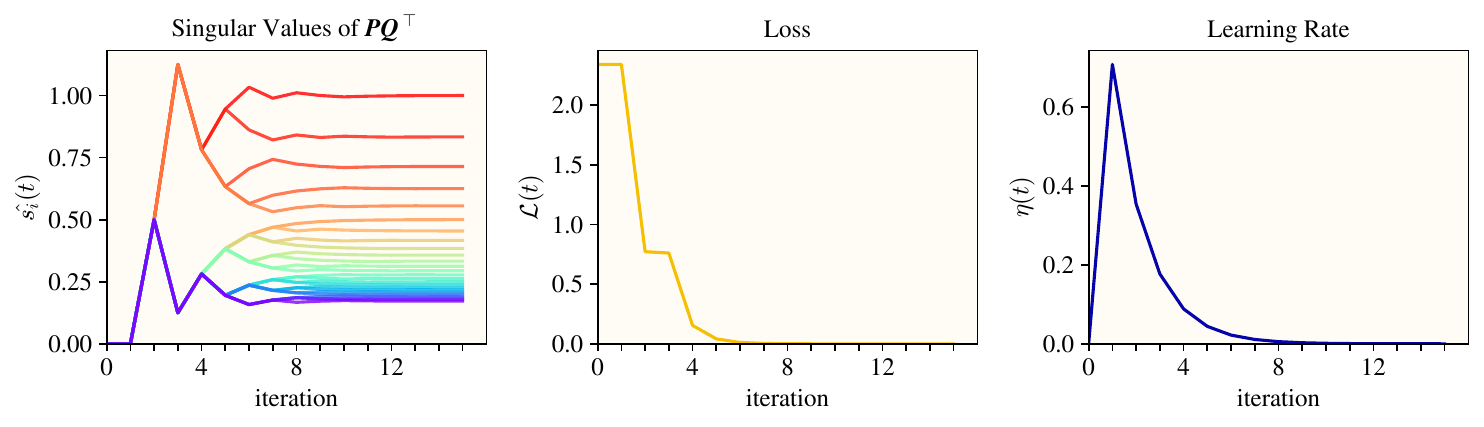}
    \caption{
    The spiked learning rate schedule in the exactly-parameterized setting ($n = d = 25$) with
    orthogonal initialization at scale $\alpha = 10^{-4}$.
    \textbf{(Left.)}~Singular values of $\mP\mQ^\top$ over $16$ iterations.
    \textbf{(Middle.)}~Loss $\mc{L}(\mP,\mQ)$.
    \textbf{(Right.)}~Learning rate schedule.
    The first step ($\eta^{(0)} = \alpha = 10^{-4}$) achieves near-perfect internal alignment.
    The second step ($\eta^{(1)} = \sqrt{s^*_{\max}/2} \approx 0.707$) simultaneously achieves
    near-perfect external alignment and grows all singular values toward $s^*_{\max}/2$.
    Geometrically annealing by $1/2$ thereafter yields rapid convergence.
    }
    \label{fig:schedule}
\end{figure}

We assume $d\geq n$ and small (semi-)orthogonal initial weights.%
\footnote{In the \textit{heavily} overparameterized setting, $d\gg n$, the following assumptions hold approximately with high probability under random Gaussian initialization, and the rest of the argument carries over as well.}
Then, at initialization, it holds that $\mP\mP^\top =  \mQ\mQ^\top = \alpha^2 \mI_n$ and $\Mstar \gg \mP\mQ^\top$.
The initial updates will be:

\begin{align}
    \ort{\nabla_\mP \mc{L}} \approx \ort{-\Mstar\mQ} &= -\frac{1}{\alpha}\mQ \\
    \ort{\nabla_\mQ \mc{L}} \approx \ort{-\Mstar\mP} &= -\frac{1}{\alpha}\mP.
\end{align}

The diagonal $\Mstar$ term in the gradient will simply rescale the already orthogonal rows of $\mP$ and $\mQ$. However, the orthogonalization step of Muon will then immediately rescale the rows again to make them unit length. Thus, at initialization, the updates are approximately: 

\begin{align}
    \mP^{(1)} &\approx \mP^{(0)} + \frac{\eta^{(0)}}{\alpha}\mQ^{(0)} \\
    \mQ^{(1)} &\approx \mQ^{(0)} + \frac{\eta^{(0)}}{\alpha}\mP^{(0)}.
\end{align}

If we set the initial learning rate $\eta^{(0)}$ equal to $\alpha$, then $\mP$ and $\mQ$ align perfectly in exactly one step: $\mP^{(1)} \approx \mP^{(0)}+\mQ^{(0)} \approx \mQ^{(1)}$. The second update will approximately be:
\begin{align}
    \mP^{(2)} &\approx \mP^{(1)}+\eta^{(1)} \ort{\Mstar\mP^{(1)}} \\
     &\approx \mQ^{(2)}
\end{align}
where we used the fact that $\Mstar \gg \mP^{(1)}(\mQ^{(1)})^\top$ (as long as the initial learning rate is small) and that $\mP^{(1)}\approx \mQ^{(1)}$. If the learning rate $\eta^{(1)}$ is sufficiently large, the update matrix $\ort{\Mstar\mP^{(1)}}$ will overwhelm $\mP^{(1)}$ and $\mQ^{(1)}$ such that $\mP^{(2)} \approx \eta^{(1)}\ort{\Mstar\mP^{(1)}} \approx \mQ^{(2)}$. As long as $n \leq d$, this forces the rows of $\mP^{(2)}$ (which are exactly the rows of $\mQ^{(2)}$) to be mutually orthogonal such that $\mP^{(2)}(\mQ^{(2)})^\top \approx (\eta^{(1)})^2 \mI_n$, thus achieving near-perfect alignment with the target. See \cref{fig:schedule} for empirical validation of our assumptions and the efficacy of our resulting schedule.
 
We are free to set the initial learning rate $\eta^{(0)}$ and the learning rate spike $\eta^{(1)}$ to any values that satisfy $\eta^{(0)} = \alpha$ and $\eta^{(1)} \gg \eta^{(0)}$. After the two steps that achieve alignment, the objective decouples into $n$ independent scalar problems; geometrically annealing the learning rate guarantees linear convergence as shown in \cref{sec:anneal}. 

\paragraph{Limitations.}
Our analysis is restricted to simple matrix factorization problems. We focus on the unmodified spectral gradient descent update rule, i.e., ignoring the effects of momentum, weight decay, and Newton-Schulz iteration. Moving beyond these simplifications is an interesting and important area for future work.

\paragraph{Acknowledgements.}
We thank Joey Turnbull and Jianhao Ma for useful discussions.
This work was funded by Imbue under the Feature Lab (FLAB) initiative.

\paragraph{Author contributions.}
JS and DK conceived the idea in the car while crossing the Bay Bridge and oversaw the project's subsequent development. MR and DK jointly performed the theoretical analysis, coded the numerical experiments, and wrote the manuscript with input and guidance from JS.

\clearpage
\bibliographystyle{tmlr}
\bibliography{references}

\appendix
\clearpage
\section{Experimental details}
\label[app]{app:expt}

All experiments solve the matrix factorization problem
$\mc{L}(\mP,\mQ) = \frac{1}{2}\|\Mstar - \mP\mQ^\top\|^2$
with two trainable factors $\mP\in\R^{n\times d}$ and $\mQ\in\R^{n\times d}$, optimized either by plain (momentum-free, no weight decay) gradient descent or by the momentum-free Muon update of \cref{sec:intro}.
The following conventions apply: the target $\Mstar$ is taken diagonal (we always work in its SVD basis, without loss of generality), the factors are initialized i.i.d.\ Gaussian with $\mP_\idx{ij},\mQ_\idx{ij}\sim\mc{N}(0,\alpha^2/\max(n,d))$ at initialization scale $\alpha$, and $\eta$ denotes the learning rate.
Our Muon implementation computes the update direction as the \emph{exact} polar factor $\ort{\mG}=\mU\mV^\top$ from a full SVD $\mG=\mU\mS\mV^\top$ of the gradient, rather than the Newton--Schulz approximation used in practice; this removes orthogonalization error as a confound.

Code is publicly available at \href{https://github.com/dkarkada/muon-mfac}{https://github.com/dkarkada/muon-mfac}.

\subsection{Summary of dynamical differences (\texorpdfstring{\cref{fig:summary}}{summary figure})}

This figure comprises three independent experiments (one per row, GD in the left column and Muon in the right), all run on the same target. The target is the diagonal $n\times n$ matrix with $n=32$ and singular values $s_\mu = 10\,\mu^{-1}$ for $\mu\in[1,\dots,n]$. The models are all exactly-parameterized, $d=n$.

\paragraph{Singular-value trajectories (panels a, b).} We train from initialization scale $\alpha=5\times10^{-3}$ for $1000$ steps, with $\eta=5\times10^{-3}$ for Muon and $\eta=1.5\times10^{-2}$ for GD. Each panel plots all $32$ singular values of the factor $\mP$ as a function of optimization time, colored from the largest mode to the smallest. The axes are unlabeled because the panels are meant to be read qualitatively. The time axis is scaled linearly and the singular value axis ranges linearly from $[0,3.7]$.

\paragraph{Final loss versus learning rate (panels c, d).} Here we sweep the learning rate to probe stability. Both axes are scaled logarithmically. We use a larger initialization scale $\alpha=0.1$ and sweep $200$ learning rates spaced logarithmically over $[0.05,\,20]\times\eta_{\mathrm c}$ with $\eta_{\mathrm c}=0.14$, a reference rate marking GD's critical (divergence) threshold. Each Muon run is $500$ steps and each GD run is $2000$ steps. The plotted ``final loss'' is the average of the last two iterates' losses (to suppress the terminal oscillation of Muon) normalized by the initial loss.  The red dashed line marks the GD $\eta_{\mathrm c}$ and the gray dashed line marks the initial loss.

\paragraph{Conserved quantities (panels e, f).} We train from $\alpha=5\times10^{-3}$ for $1200$ steps, with $\eta=5\times10^{-4}$ for Muon and $\eta=2.5\times10^{-3}$ for GD, and track the two candidate conserved matrices
$\mDelta_1 \defn \sqrt{\mP^\top\mP}-\sqrt{\mQ^\top\mQ}$ (Muon) and
$\mDelta_2 \defn \mP^\top\mP-\mQ^\top\mQ$ (GD).
To test conservation we fix a reference iterate $t^\ast=80$ (chosen after alignment, during the growth phase) and plot, for each $\Delta\in\{\mDelta_1,\mDelta_2\}$, the overlap $\langle\Delta(t),\,\Delta(t^\ast)\rangle_\mathrm{F}/\|\Delta(t^\ast)\|_\mathrm{F}^2$. This overlap equals $1$ exactly when $\Delta(t)=\Delta(t^\ast)$, so a curve pinned at $1$ for $t>t^\ast$ indicates that quantity is conserved. Under GD (panel e) it is $\mDelta_2$ that stays fixed; under Muon (panel f) it is $\mDelta_1$.

\clearpage

\subsection{Alignment rates (\texorpdfstring{\cref{fig:fig3}}{Figure 3})}
\label{app:alignment}

This experiment measures how quickly the weights align, as a function of the aspect ratio $d/n$ and the initialization scale $\alpha$, and compares the measured rates to the power laws derived in \cref{sec:alignment}. All runs use Muon (no GD comparison here) in \texttt{float64} for numerical precision, and each configuration is averaged over $3$ random seeds.

\paragraph{Target.} We fix $n=128$ and build a planted-signal-in-noise target: form $\mM = \mathrm{diag}(\lambda\,\mathbf{1}_{r}) + \mG/\sqrt{n}$ with planted rank $r=32$, planted signal strength $\lambda=4$, and $\mG$ i.i.d.\ standard Gaussian; the target singular values are then $10\times$ the singular values of $\mM$, floored at $2$ to keep the noise bulk well-conditioned. Equivalently $\Mstar=\mG'+\lambda\,\mU\mV^\top$ with $\mU,\mV\in\R^{n\times32}$ semi-orthogonal, as stated in the caption. As elsewhere we rotate into the target's SVD basis, so $\Mstar$ is diagonal. The relative spectral gap $s$ between the planted and bulk singular values sets the under-parameterized alignment exponent $2(1-s^{-1})$; we compute its effective value by taking the mean relative gap between signal modes and the bulk edge.

\paragraph{Sweep.} We vary the aspect ratio over $d\in\{32,128,512\}$, i.e.\ $d/n\in\{0.25,1,4\}$ (under-, exactly-, and over-parameterized), and the initialization scale over $\alpha\in\{10^{-5},10^{-4},10^{-3}\}$, with learning rate $\eta=0.2\,\alpha$ held small relative to $\alpha$.

\paragraph{Metrics and axes.} We report the misalignments $1-a_\mathrm{in}$, $1-a_\mathrm{left}$, $1-a_\mathrm{right}$ using the degeneracy-aware alignment metric $a(\cdot)$ of \cref{sec:alignment}. The degenerate-subspace grouping merges adjacent modes $i,j$ unless \emph{both} their relative singular-value gap exceeds a tolerance (here $20\%$, with a small absolute floor) in \emph{both} the model and the reference spectra; this is the convention referenced in \cref{sec:alignment}. The horizontal axis is rescaled optimization time $t/t_\mathrm{converge}$, where $t_\mathrm{converge}$ is the step at which uniform growth would carry the top mode to its target.

\paragraph{Panels.} The three columns correspond to the three aspect ratios. In the under-parameterized column ($d/n=0.25$) internal alignment is essentially free ($1-a_\mathrm{in}\approx0$), so we plot target misalignment $1-a_\mathrm{left}$ (dotted) and $1-a_\mathrm{right}$ (dashed), overlaid with the theoretical power law $t^{-2(1-s^{-1})}$. In the exactly- and over-parameterized columns ($d/n=1,4$) target alignment is essentially free, so we plot internal misalignment $1-a_\mathrm{in}$; the over-parameterized column is overlaid with the predicted $t^{-4}$ law. Within each panel the three colors correspond to the three initialization scales $\alpha$, illustrating that smaller $\alpha$ (equivalently smaller $\alpha_\mathrm{eff}=\max(\eta,\alpha)$) aligns earlier in rescaled time.

\subsubsection{Alignment rates for a general target spectrum}
\label{app:general-spectrum}

\Cref{fig:fig3} used a planted-signal-in-noise target because its spectrum has a single well-separated gap (between the planted modes and the noise bulk), letting us summarize the under-parameterized alignment rate with one effective gap $s$. Here we check that the mechanism survives for a target with no such special structure: a smooth power-law spectrum.

\paragraph{Target.} We fix $n=64$ and set $\sigma_\mu^\ast = 10\,\mu^{-p}$ for $\mu\in\{1,\dots,n\}$, with $p=1$. (Larger $p$ widens the relative gap at the model-capacity edge, recovering the large-gap regime of \cref{fig:fig3}; $p=1$ is a generic exponent for which this gap is small.)

\paragraph{Sweep.} As in \cref{fig:fig3}, we run Muon in \texttt{float64} at $\alpha\in\{10^{-4},10^{-5},10^{-6}\}$ with learning rate kept small relative to $\alpha$, and consider an under-parameterized ($d=16$, $d/n=0.25$) and an over-parameterized ($d=256$, $d/n=4$) aspect ratio. The under-parameterized runs average $500$ seeds per $\alpha$; the over-parameterized runs average $20$.

\paragraph{Metrics and axes.} We report the misalignments $1-a_\mathrm{in}$, $1-a_\mathrm{left}$, $1-a_\mathrm{right}$ using the degeneracy-aware alignment metric $a(\cdot)$ of \cref{sec:alignment} as in \cref{fig:fig3}. The horizontal axis is rescaled optimization time $t/t_\mathrm{converge}$, where $t_\mathrm{converge}$ is the step at which uniform growth would carry the top mode to its target.

\begin{figure}[t]
\centering
\begin{minipage}{0.48\textwidth}
\centering
\includegraphics[width=\linewidth]{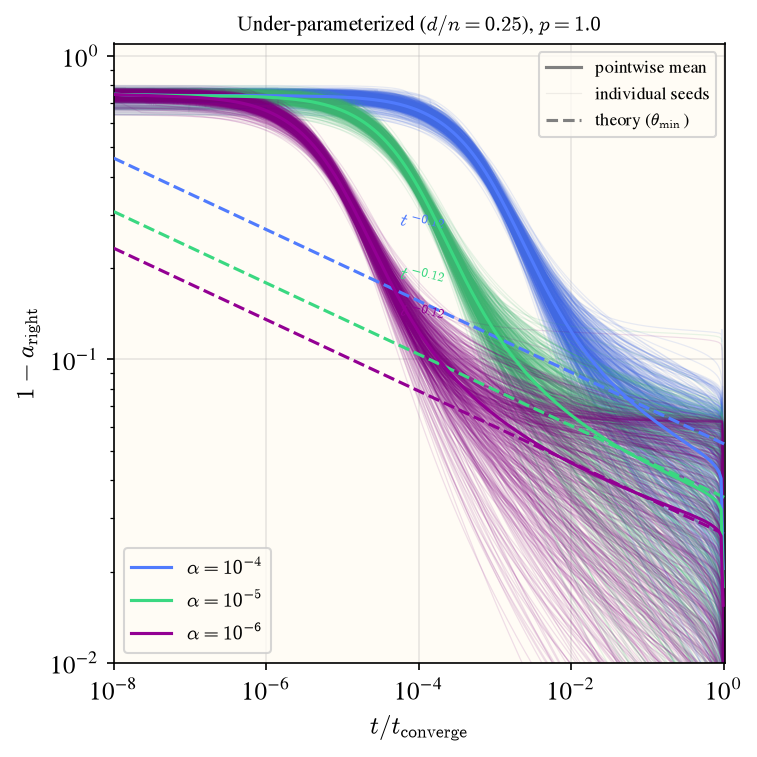}
\end{minipage}
\hfill
\begin{minipage}{0.48\textwidth}
\centering
\includegraphics[width=\linewidth]{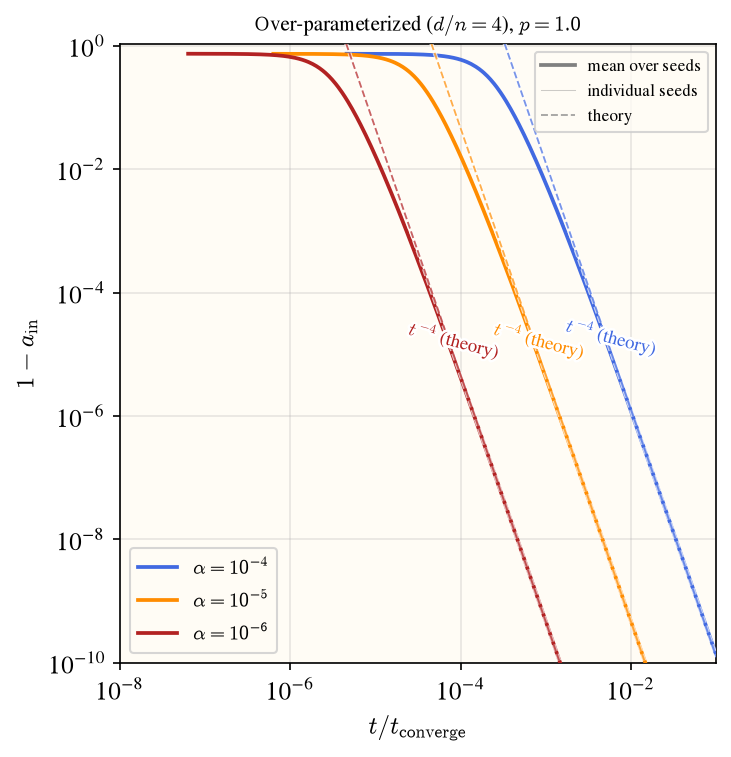}
\end{minipage}
\caption{Alignment rates for the target with power-law spectrum ($p=1.0$). Left: under-parameterized ($d/n=0.25$), $1-a_\mathrm{right}$ vs.\ $t/t_\mathrm{converge}$, overlaid with the edge-mode power law $t^{-2(1-s_\mathrm{edge}^{-1})}$. Right: over-parameterized ($d/n=4$), $1-a_\mathrm{in}$ vs.\ $t/t_\mathrm{converge}$, overlaid with the universal $t^{-4}$ law. Colors denote initialization scale $\alpha\in\{10^{-4},10^{-5},10^{-6}\}$.}
\label{fig:general-spectrum}
\end{figure}

\paragraph{Panels.} In the under-parameterized panel (\cref{fig:general-spectrum}, left) we generalize \cref{sec:outer-prod} by applying its single-gap result locally at the model-capacity edge: with $s_\mathrm{edge}\defn\sigma_d^\ast/\sigma_{d+1}^\ast$ the relative gap between the last mode the model can represent and the first it cannot, the tail of $1-a_\mathrm{right}$ is well-approximated by the predicted alignment rate of the slowest mode $t^{-2(1-s_\mathrm{edge}^{-1})}$. Intuitively, each mode aligns at its own gap-dependent rate, so once the faster (larger-gap) modes have essentially aligned, whatever misalignment remains is increasingly dominated by the one mode that aligns slowest---i.e., the edge mode---making its rate the natural candidate for the asymptotic tail. Since \cref{sec:outer-prod}'s rate is itself derived under a small-initialization approximation, the theory predicts this match should sharpen as $\alpha$ shrinks; averaged over seeds, the measured tail indeed tracks the predicted rate more closely as $\alpha$ decreases across the swept values. We make no claim about the shape of the curve before the tail, which need not follow a single power law. In the over-parameterized panel (\cref{fig:general-spectrum}, right), $1-a_\mathrm{in}$ closely follows the universal, target-independent rate $t^{-4}$ of \cref{sec:inner-prod} at all three $\alpha$, as expected since this rate does not depend on the target spectrum. Together with \cref{fig:fig3}, this shows that (at least) the tail of alignment---the part that determines whether the process stalls short of completion---continues to decay as a power law even for a target spectrum with no special large-gap structure.

\clearpage
\subsection{Spiked learning rate schedule (\texorpdfstring{\cref{fig:schedule}}{Figure 4})}

This experiment demonstrates the spiked learning rate schedule of \cref{sec:schedule} in the
exactly-parameterized setting and verifies that alignment and convergence can be achieved in a
small number of steps.

\paragraph{Setup.}
We take $n = d = 25$ and a diagonal target $\Mstar \in \R^{n \times n}$ with spectrum
$s^*_\mu = (4+\mu)^{-1}$ for $\mu \in [1,\dots,n]$, normalized so that $s^*_1 = 1$.
Factors $\mP, \mQ \in \R^{n \times d}$ are initialized orthogonally: we draw
$\mW_1, \mW_2 \sim \mc{N}(0,1)^{n\times d}$, compute their QR factorizations
$\mW_k = \mQ_k \mR_k$, and set
$\mP^{(0)} = \alpha\,\mQ_1\,\mathrm{diag}(\mathrm{sign}(\mathrm{diag}(\mR_1)))$
(and analogously for $\mQ^{(0)}$), where the sign correction makes the initialization unique.
The initialization scale is $\alpha = 10^{-4}$.

\paragraph{Schedule.}
We run $T = 16$ steps of the momentum-free Muon update (exact polar factor via full SVD) with the
following three-phase learning rate schedule:
\begin{enumerate}
    \item \textbf{Alignment step} ($t=0$): $\eta^{(0)} = \alpha = 10^{-4}$.  As argued in
          \cref{sec:schedule}, setting $\eta^{(0)} = \alpha$ achieves near-perfect internal
          alignment in a single step.
    \item \textbf{Spike} ($t=1$): $\eta^{(1)} = \sqrt{s^*_{\max}/2} = \sqrt{1/2} \approx 0.707$.
          This large step simultaneously achieves near-perfect external (target) alignment and
          uniformly grows the singular values of $\mP\mQ^\top$ toward $s^*_{\max}/2$.
    \item \textbf{Geometric anneal} ($t \geq 2$): $\eta^{(t)} = \eta^{(1)} \cdot 2^{-(t-1)}$.
          As shown in \cref{sec:anneal}, geometric annealing guarantees linear convergence once
          the iterates are close to the solution manifold.
\end{enumerate}

\paragraph{Diagnostics.}
We track three quantities at each step: (i)~the singular values of $\mP^{(t)}(\mQ^{(t)})^\top$,
(ii)~the loss $\mc{L}(\mP^{(t)},\mQ^{(t)})$, and (iii)~the learning rate $\eta^{(t)}$, plotted
against iteration on a linear axis (panels a, b, c respectively).

\begin{figure}[t]
\centering
\renewcommand{\arraystretch}{1.3}
\begin{tabular}{@{}l p{0.27\textwidth} p{0.27\textwidth} p{0.27\textwidth}@{}}
\toprule
 & \multicolumn{1}{c}{\textbf{Full-batch Muon}} & \multicolumn{1}{c}{\textbf{Minibatch Muon}} & \multicolumn{1}{c}{\textbf{Minibatch Muon w/ Momentum}} \\
 & \multicolumn{1}{c}{\scriptsize (exact gradient)} & \multicolumn{1}{c}{\scriptsize $B=4$} & \multicolumn{1}{c}{\scriptsize $B=4$,\ $\mu=0.9$} \\
\midrule
&
\begin{center}\vspace{-6pt}\includegraphics[width=0.8\linewidth]{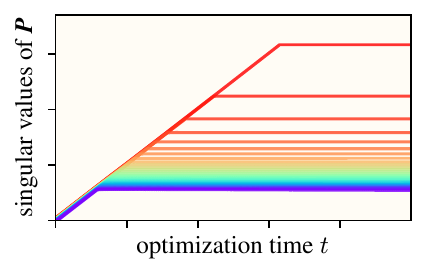}\vspace{-4pt}\end{center}
&
\begin{center}\vspace{-6pt}\includegraphics[width=0.8\linewidth]{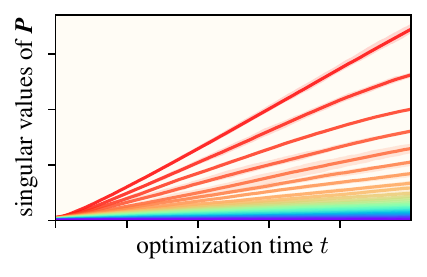}\vspace{-4pt}\end{center}
&
\begin{center}\vspace{-6pt}\includegraphics[width=0.8\linewidth]{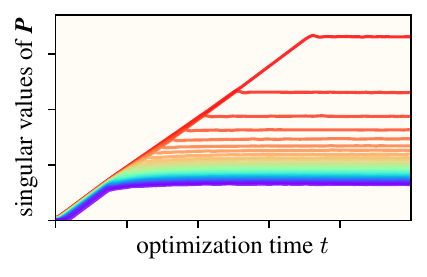}\vspace{-4pt}\end{center}
\\
\midrule
&
\begin{center}\vspace{-6pt}\includegraphics[width=0.8\linewidth]{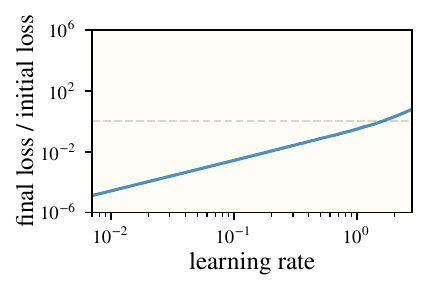}\vspace{-4pt}\end{center}
&
\begin{center}\vspace{-6pt}\includegraphics[width=0.8\linewidth]{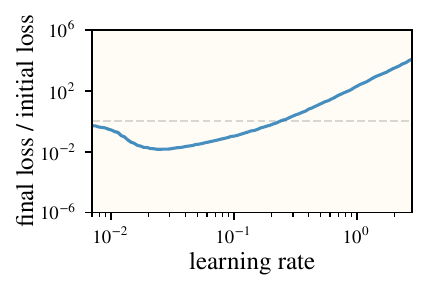}\vspace{-4pt}\end{center}
&
\begin{center}\vspace{-6pt}\includegraphics[width=0.8\linewidth]{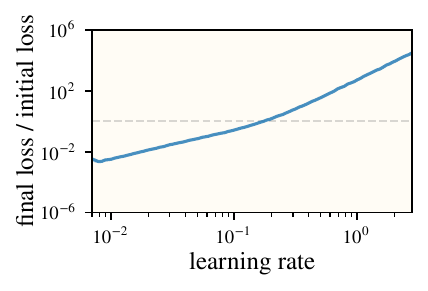}\vspace{-4pt}\end{center}
\\
\midrule
&
\begin{center}\vspace{-6pt}\includegraphics[width=0.8\linewidth]{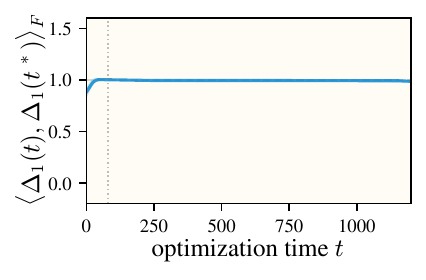}\vspace{-4pt}\end{center}
&
\begin{center}\vspace{-6pt}\includegraphics[width=0.8\linewidth]{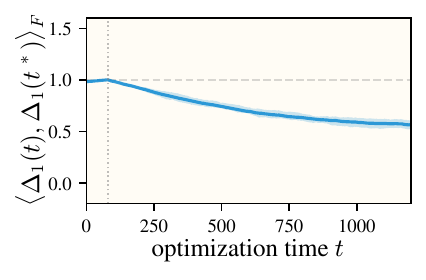}\vspace{-4pt}\end{center}
&
\begin{center}\vspace{-6pt}\includegraphics[width=0.8\linewidth]{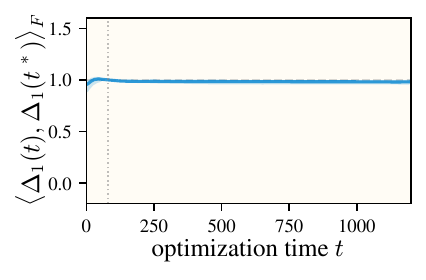}\vspace{-4pt}\end{center}
\\
\bottomrule
\end{tabular}
\caption{Robustness of \cref{fig:summary}'s Muon column to minibatch noise (Gaussian-sensing, $B=4$) and heavy-ball momentum ($\mu=0.9$), averaged over $5$ seeds with shared per-seed initializations across columns; shading shows the min--max range. The loss can never diverge regardless of the batch size or the existence of momentum. Momentum restores the deterministic singular-value staircase (row 1) and the $\mDelta_1$ conservation law (row 3).}
\label{fig:stoch}
\end{figure}

\clearpage
\section{Robustness to minibatch noise and momentum (\texorpdfstring{\cref{fig:stoch}}{stochasticity figure})}
\label[app]{app:stoch}

The comparisons in \cref{fig:summary} use exact (full-batch) gradients and no momentum, whereas Muon is used in practice with both minibatch gradients and heavy-ball or Nesterov momentum \parencite{jordan2024}. Here we check which of the three dynamical phenomena from \cref{fig:summary}'s Muon column survive under stochasticity, and whether momentum restores them.

Stochastic Muon keeps the exact polar factor but replaces the gradient by an online matrix-sensing estimate and adds heavy-ball momentum: at each step we draw a fresh batch of $B=4$ i.i.d.\ Gaussian probes $\mA_i\in\R^{n\times m}$, observe $y_i=\langle\mA_i,\Mstar\rangle$, and form
\begin{equation*}
    \mG = \frac1B\sum_{i=1}^B \langle \mA_i, \mP\mQ^\top-\Mstar\rangle\,\mA_i,
    \qquad
    \mB_\mP \leftarrow 0.9\mB_\mP + \mG\mQ, \quad \mB_\mQ \leftarrow 0.9\mB_\mQ + \mG^\top\mP,
\end{equation*}
and update along $\ort{\mB_\mP}$ and $\ort{\mB_\mQ}$.
The estimator is unbiased, $\E{}{\mG}=\mP\mQ^\top-\Mstar$. 

\paragraph{Setup.} All three columns of \cref{fig:stoch} reuse the exact target, architecture, and per-row hyperparameters ($\alpha$, $\eta$, step count) of the corresponding row in \cref{fig:summary}'s Muon column (see \cref{app:expt} above), so the deterministic column is a direct re-run of that baseline. Batch size $B=4$ and momentum $\mu=0.9$ were chosen from a preliminary sweep over $B\in\{4,16,64\}$ and $\mu\in\{0,0.9,0.99\}$: $B=4$ is small enough to visibly break the deterministic phenomena without momentum, and $\mu=0.9$ recovers them cleanly, whereas $\mu=0.99$ introduces an oscillatory overshoot artifact. All panels average over $5$ random seeds; for each seed, the three columns share an identical initialization, so differences between columns reflect only the training dynamics, not initialization variance. Shaded bands show the min--max range across seeds.

\paragraph{Singular-value trajectories (row 1).} As in \cref{fig:summary} panel b ($\alpha=5\times10^{-3}$, $\eta=5\times10^{-3}$, $1000$ steps), we plot all $32$ singular values of $\mP$ against optimization time. Minibatch noise without momentum does not merely add variance to the deterministic staircase: it systematically suppresses the smaller, slower-growing modes, which stall near zero instead of eventually growing in (final singular values of the bottom $8$ modes average $0.05$, versus $0.61$ in the deterministic run). Momentum restores a staircase close to the deterministic one (bottom-$8$ average $0.70$).

\paragraph{Maximum stable learning rate (row 2).} As in \cref{fig:summary} panel d ($\alpha=0.1$, $500$ steps, learning rate swept logarithmically over $[0.05,20]\times\eta_{\mathrm c}$ with $\eta_{\mathrm c}=0.14$), we plot final loss (normalized by initial loss, averaged over the last $20$ iterates) against learning rate on log--log axes. The fact that the loss cannot diverge is unaffected by noise or momentum since $\|\mathrm{orth}(\cdot)\|_{\mathrm{op}}=1$ regardless of its argument. 

\paragraph{Conserved quantities (row 3).} As in \cref{fig:summary} panel f ($\alpha=5\times10^{-3}$, $\eta=5\times10^{-4}$, $1200$ steps, reference iterate $t^\ast=80$), we track the overlap $\langle\mDelta_1(t),\mDelta_1(t^\ast)\rangle_{\mathrm F}/\|\mDelta_1(t^\ast)\|_{\mathrm F}^2$. Under stochasticity without momentum this overlap does not merely fluctuate around $1$; it drifts systematically and monotonically away from conservation over the course of training ($0.92$ at $t=200$, $0.70$ at $t=600$, $0.56$ at $t=1199$), reflecting that $\E{}{\ort{\mG+\epsilon}} \ne \ort{\E{}{\mG+\epsilon}}$ in general, so the bias compounds step by step. Momentum almost fully restores conservation ($\geq 0.98$ throughout), matching the deterministic column within seed variance.

\clearpage
\section{Muon reaches balanced solutions in practical settings with vanishing initialization}
\label[app]{app:bigsweep}

A central claim of this paper is that Muon, like GD, reaches increasingly \emph{balanced} solutions as the initialization scale vanishes.
The argument is given in \cref{sec:scalarfac}: once the factors have aligned, the Muon flow conserves $\mDelta_1 \defn \sqrt{\mP^\top\mP}-\sqrt{\mQ^\top\mQ}$, whose norm is $O(\alpha)$ at initialization, so the \emph{balance defect} $\|\mP^\top\mP-\mQ^\top\mQ\|_F^2$ at convergence should vanish as $\alpha^2$.
That argument assumes an idealized setting: exact orthogonalization, exact gradients, a square target, and a learning rate small enough that the discrete iteration tracks its flow.
Muon as it is actually used satisfies none of the first three.
We therefore ran a sweep that relaxes each idealization in turn and measures the balance defect directly.

\paragraph{Problem instances.} We again minimize a matrix factorization problem, but with potentially rectangular target $\Mstar\in\R^{n\times m}$. We set $\Mstar$ by drawing independent Haar-random semi-orthogonal $\mU^\ast\in\R^{n\times r}$ and $\mV^\ast\in\R^{m\times r}$ with $r=\min(n,m)$ and set $\Mstar=\mU^\ast\mS^\ast{\mV^\ast}^\top$ with spectrum $s^\ast_\mu=50\,\mu^{-\gamma}$, i.e., power-law with exponent $\gamma$.
The left and right singular bases are random, so $\Mstar$ is neither symmetric nor aligned with the initialization in general.
We probe three shapes: exactly-parameterized ($(n,m,d)=(50,50,50)$), under-parameterized ($(n,m,d)=(50,50,10)$), and rectangular target ($(n,m,d)=(50,20,20)$).
We probe two target exponents, $\gamma\in\{1,\,0.2\}$ (a fast-decaying and a nearly-flat spectrum).

\paragraph{Optimizers.} We probe three variants of Muon.
\emph{Idealized Muon} uses the exact polar factor from a full SVD and the exact full-batch gradient, as in the rest of the paper.
\emph{Newton--Schulz Muon} replaces the exact polar factor by $10$ steps of the classical cubic iteration $\mX \leftarrow \frac32\mX-\frac12\mX\mX^\top\mX$ on the Frobenius-normalized gradient. This reflects how Muon is implemented in practice, which typically leaves the tail of the spectrum under-orthogonalized.
\emph{Stochastic Muon} keeps the exact polar factor but replaces the gradient by an online matrix-sensing estimate and adds heavy-ball momentum: at each step we draw a fresh batch of $B=64$ i.i.d.\ Gaussian probes $\mA_i\in\R^{n\times m}$, observe $y_i=\langle\mA_i,\Mstar\rangle$, and form
\begin{equation*}
    \mG = \frac1B\sum_{i=1}^B \langle \mA_i, \mP\mQ^\top-\Mstar\rangle\,\mA_i,
    \qquad
    \mB_\mP \leftarrow 0.9\mB_\mP + \mG\mQ, \quad \mB_\mQ \leftarrow 0.9\mB_\mQ + \mG^\top\mP,
\end{equation*}
and update along $\ort{\mB_\mP}$ and $\ort{\mB_\mQ}$.
The estimator is unbiased, $\E{}{\mG}=\mP\mQ^\top-\Mstar$.
We always report the \emph{full} reconstruction loss and not the sensing loss.

\paragraph{Initialization and learning rate.} The initialization is the same as elsewhere in the paper, with the scale normalizer taken from each factor's own shape so that the rectangular configuration is not imbalanced at initialization by construction.
We sweep four initialization scales $\alpha\in\{10^{-5},10^{-4},10^{-3},10^{-2}\}$ with $5$ seeds each; the seed fixes both the target and the direction of the initialization, so trials are paired across $\alpha$ and across optimizers.
The learning rate schedule is such that $\eta$ is 10 times smaller than the smallest singular value of $\mP$ and $\mQ$; this way, the discrete trajectory tracks the flow \emph{at every $\alpha$}. After a fixed flow time (set so that the model is very near the minimizer) we anneal the learning rate geometrically, $\eta\leftarrow0.9\,\eta$ for $300$ steps, enabling convergence.

\begin{figure}[t]
\centering
\includegraphics[width=\textwidth]{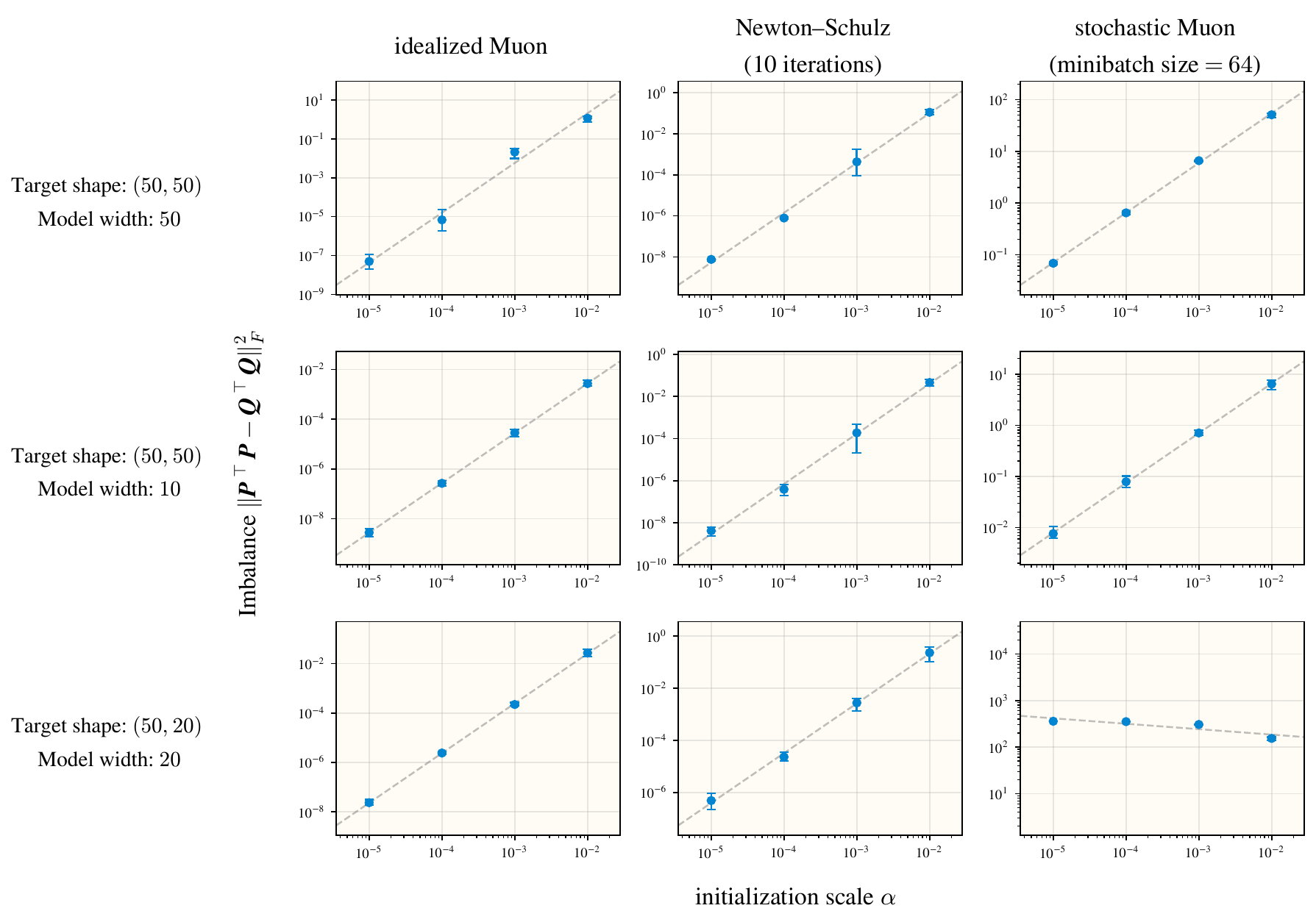}
\caption{\textbf{Muon reaches increasingly balanced solutions from vanishing initialization, in every setting but one.}
Each panel plots the balance defect $\|\mP^\top\mP-\mQ^\top\mQ\|_F^2$ at the end of training against the initialization scale $\alpha$, on log--log axes.
Markers are means over $5$ seeds, error bars span the min--max range across seeds, and the faint dashed line is a least-squares fit in log space.
Rows vary the shape of the target matrix (which is chosen randomly with a power-law spectrum) and the model.
Columns vary the optimizer: idealized Muon (exact polar factor, full-batch gradient), Muon with $10$ Newton--Schulz iterations in place of the exact polar factor, and stochastic Muon (online Gaussian matrix sensing with heavy-ball momentum).
The loss always reaches its expressivity floor, including in the stochastic + rectangular target setting.}
\label{fig:bigsweep}
\end{figure}

\paragraph{Result.} The claim is supported in every setting except stochastic Muon + rectangular target.
\Cref{fig:bigsweep} shows the $\gamma=1$ panels: in eight of the nine, the balance defect falls monotonically as a clean power law in $\alpha$.
Every run reaches the loss floor set by the model expressivity, up to machine precision.
We find that approximate orthogonalization is harmless: $10$ Newton--Schulz iterations behave like exact orthogonalization as far as balancedness is concerned.
Under stochastic Muon with a square target the defect still vanishes, but more slowly.
The same picture holds across target spectra, so we show only $\gamma=1$ here.

However, when stochastic Muon learns a rectangular target, the defect does not decrease at all, instead reaching a large and approximately $\alpha$-independent value.
This is not a failure to optimize: the loss in these runs reaches the same floor as the others, and the singular values of $\mP\mQ^\top$ rise smoothly and saturate on their targets. It is the individual factors that end up unequal, with $\mQ$'s spectrum systematically overshooting and $\mP$'s undershooting. We conjecture the following mechanism, which our experiments support.
With $n\neq m$ the two factors see gradients of different shapes, $\mG\mQ\in\R^{n\times d}$ and $\mG^\top\mP\in\R^{m\times d}$.
The two carry identical signal, but the noise in $\mG^\top\mP$ is contracted onto $\mP$'s $d$ columns from an $n$-dimensional row space while the noise in $\mG\mQ$ is merely rotated, so the two factors' signal-to-noise ratios differ by a factor of $\sqrt{m/n}$. The SNR is thus set by the shape of the weight matrix, and momentum cannot equilibrate them.
Noise rotates the polar factor away from its noiseless counterpart and shrinks each factor's spectral speed. In other words, under the flow both factors' nuclear norms grow at exactly rate $d$, which is what conserves the Hamiltonian $H=\|\mP\|_\ast-\|\mQ\|_\ast=\Tr\mDelta_1$; under noise they grow at $\rho_\mP d$ and $\rho_\mQ d$ instead, where $\rho\leq1$ is determined by the factor's SNR.
When $n=m$ then $\rho_\mP = \rho_\mQ$ and $H$ remains conserved in expectation, but when $n\neq m$ we expect $H$ to drift at a fixed sign and at an $O(1)$ rate independent of $\eta$ and $\alpha$.

We believe this is a genuine property of stochastic Muon rather than an artifact of the experiment: the balance invariant is conserved under minibatch noise only when the two factors see gradients of the same shape.
This may be an interesting avenue for future work.

\end{document}